\pdfoutput=1

\documentclass[11pt]{article}

\usepackage[]{acl}

\usepackage{times}
\usepackage{latexsym}

\usepackage[T1]{fontenc}

\usepackage[utf8]{inputenc}

\usepackage{microtype}

\usepackage{graphicx}
\usepackage{amsmath}
\usepackage{amsfonts}
\usepackage{amsbsy}
\usepackage{hyperref}
\usepackage{bm}

\title{Modeling Content-Emotion Duality via Disentanglement for Empathetic Conversation}

\author{Peiqin Lin$^1$\thanks{Work done at The Hong Kong Polytechnic University.}, Jiashuo Wang$^2$, Hinrich Schütze$^1$, Wenjie Li$^2$ \\
        $^1$CIS \& MCML, LMU Munich, Germany\\
        $^2$Department of Computing, The Hong Kong Polytechnic University, Hong Kong\\
        \texttt{linpq@cis.lmu.de,jessiejs.wang@connect.polyu.hk,cswjli@comp.polyu.edu.hk}
}

\begin{document}
\maketitle
\begin{abstract}
The task of empathetic response generation aims to understand what feelings a speaker expresses on his/her experiences and then reply to the speaker appropriately. To solve the task, it is essential to model the content-emotion duality of a dialogue, which is composed of 
the content view (i.e., what personal experiences are described) and the emotion view (i.e., the feelings of the speaker on these experiences).
To this end, we design a framework to model the Content-Emotion Duality (CEDual) via disentanglement for empathetic response generation.
With disentanglement, we encode the dialogue history from both the content and emotion views, and then generate the empathetic response based on the disentangled representations, thereby both the content and emotion information of the dialogue history can be embedded in the generated response.
The experiments on the benchmark dataset EMPATHETICDIALOGUES show that the CEDual model achieves state-of-the-art performance on both automatic and human metrics, and it also generates more empathetic responses than previous methods.\footnote{Code is available at \href{https://github.com/lpq29743/CEDual}{https://github.com/lpq29743/CEDual}.}
\end{abstract}

\section{Introduction} \label{sec:intro}

Empathy, the capacity to understand the feelings of people on their described experiences \cite{rothschild2006help,read2019typology}, is a desirable trait in human-facing dialogue systems \cite{rashkin2018towards}. In this paper, we focus on the task of empathetic response generation, which aims to understand the feelings of the speaker as well as how the feelings emerge from the described experiences, and then generate the empathetic response.

\begin{figure}[htbp]
  \centering
  \resizebox {\columnwidth} {!} {
    \includegraphics[clip, trim=9.75cm 4.6cm 10.5cm 6.95cm]{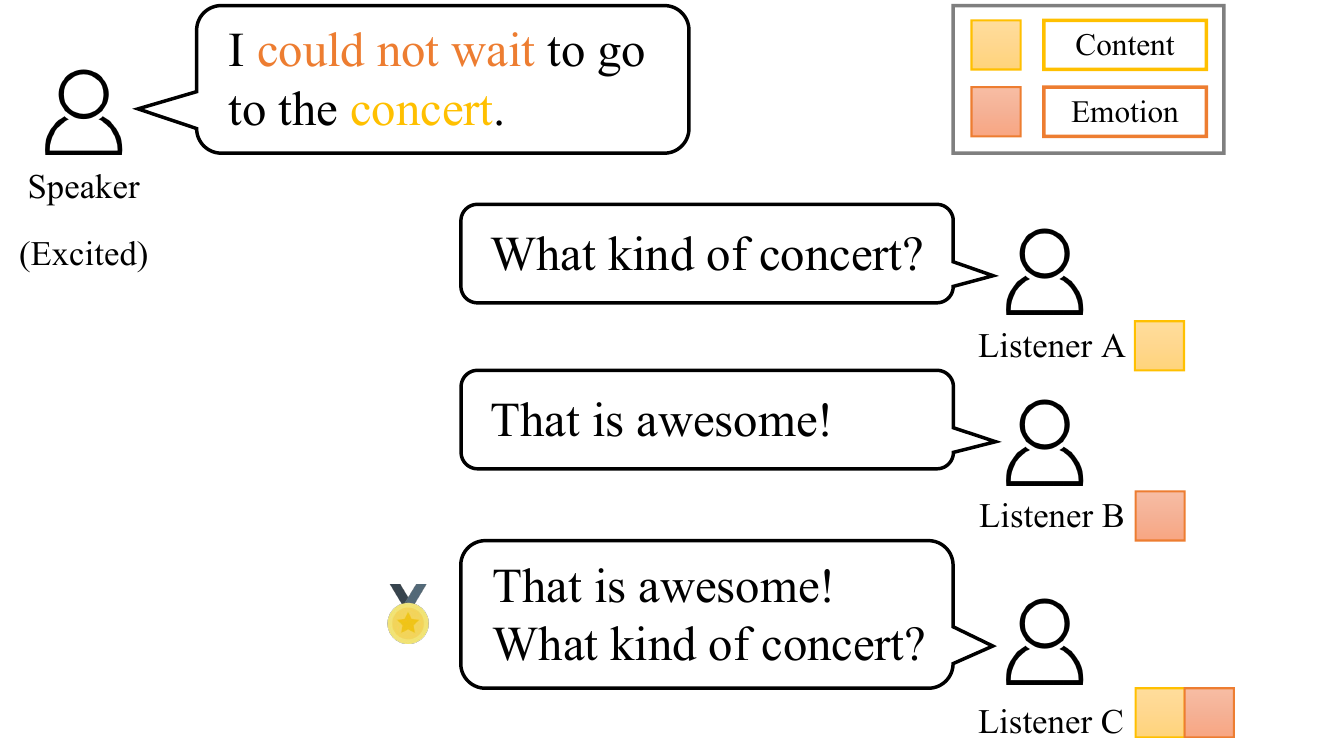}
  }
  \caption{An example of Empathetic Response Generation. ``Listener C'' provides the best response since it achnowledge the ``Speaker'' from both the content and emotion views.}
  \label{fig:example}
\end{figure}

Empathetic reflection involves paying attention to the content-emotion duality of the dialogue, which is composed of a content component and an emotion component \cite{marathe2021empathetic}. Specifically, the content component is the actual incident devoid of any feelings, while the emotion component is the feelings evoked. For example, as shown in Fig.~\ref{fig:example}, the utterance ``I could not wait to go to the concert'' from the speaker involves the content component ``concert'' and the emotion component ``could not wait'', which indicates the expressed ``excited'' emotion of the speaker. Among the responses from the listeners, ``Listener A'' focuses on the content component alone, while ``Listener B'' just focuses on the emotion component. Neither Listener A nor B considers both the content and emotion components, thus failing to acknowledge the speaker on both the feelings of the speaker and the facts where the feelings emerge. An empathetic listener, like ``Listener C'', is required to generate the response, which has high correlations with not only the content component but also the emotion component of the speaker utterance.

In real-world human cognitive processes, emotion is completely separate from content, such as facts or incidents \cite{pettinelli2012psychology,scarantino2018emotion}. Taking Fig.~\ref{fig:example} as an example, the content component ``concert'' can evoke different feelings, while the emotion component ``could not wait (excited)'' can also be caused by different incidents. Therefore, to model the content-emotion duality of the empathetic conversation, it is essential to disentangle the representation of the dialogue context onto the content space and the emotion space to better understand the dialogue context. However, the previous methods \cite{rashkin2018towards,lin2019moel,li2019empdg} encoded the content and emotion information of the speaker with the same entangled representation, which weakens the capacity of the models to effectively capture what the content and emotion information are expressed in the dialogue history.

\begin{figure}
  \centering
  \resizebox {\columnwidth} {!} {
    \includegraphics[clip, trim=10cm 4.5cm 10cm 4.5cm]{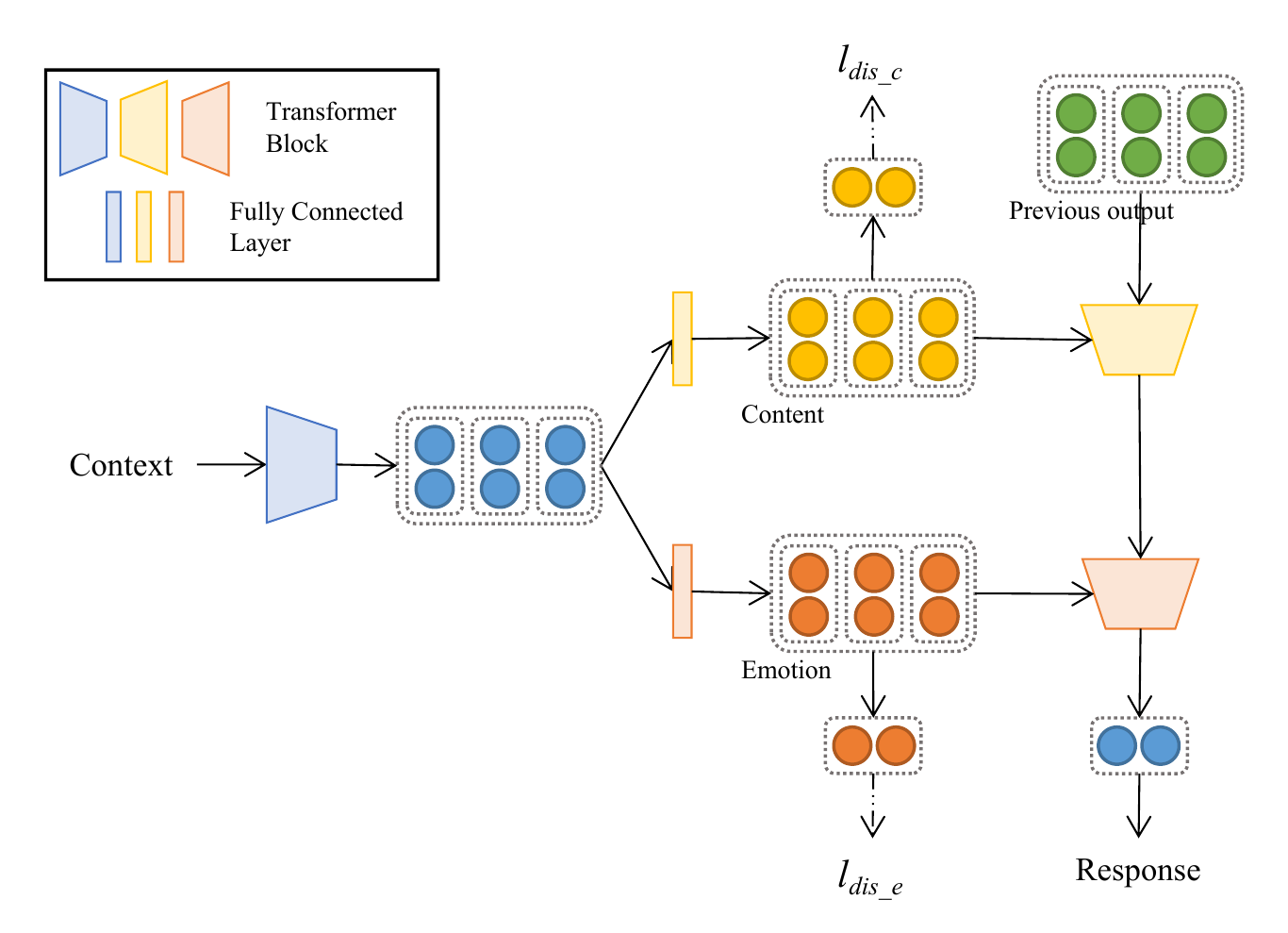}
  }
  \caption{CEDual with the first-content-then-emotion decoder.}
  \label{fig:model}
\end{figure}

To address the above-mentioned issue, we propose a framework to model the Content-Emotion Duality (CEDual) of the dialogue via disentanglement, as shown in Fig.~\ref{fig:model}, for empathetic response generation. In the proposed CEDual, the representation of the history context is disentangled onto the content space and the emotion space with two auxiliary constraints based on the emotion label. Using the disentangled content-aware and emotion-aware features, we propose two methods, namely, the first-content-then-emotion method (CEDual-FCTE) and the first-emotion-then-content method (CEDual-FETC), to imitate empathetic reflection step by step.
To examine the effectiveness of the proposed framework, we conduct experiments on the benchmark dataset EMPATHETICDIALOGUES \cite{rashkin2018towards}. The results show that our model achieves state-of-the-art performance.


\section{Related Work}

Early approaches \cite{zhou2017mojitalk,zhou2018emotional,huang2018automatic,colombo2019affect,song2019generating,shen2020cdl} focus on the emotion controllable generation to build empathetic conversational agents. Given the dialogue history and the specific emotion label, the model is required to generate the response where the desired emotion is expressed. Specifically, these methods encode the given emotion category as a vector and then add it to the decoding process for generating the emotion-aware response. However, they consider the emotion information in a hard-coded manner, thus ignoring the emotion expressed in the dialogue history.

To alleviate the above problem, some researchers \cite{li2018syntactically,rashkin2018towards} began to focus on identifying the emotion information expressed by the speaker, and then generating the response based on the identified emotional information. \citet{li2018syntactically} predict the emotion and topic keywords that should appear in the final reply and then generate the reply based on the predicted keywords. \citet{rashkin2018towards} release a large-scale dataset, namely EMPATHETICDIALOGUES, and propose a benchmark model, which adopts an external emotion classifier to identify the emotion expressed by the speaker and then generate the empathetic response.

Following \citet{rashkin2018towards}, \citet{lin2019moel} softly combine the possible emotional responses from several separate decoders to generate the final empathetic response; \citet{li2019empdg} introduce word-level emotional information to better perceive the emotion of the dialogue history and further consider the effect of user feedback via a novel interactive adversarial mechanism; \citet{DBLP:journals/kbs/WangLLM21} propose a graph-based network to reason emotional causality for empathetic response generation. Although promising results are achieved by the above approaches, they represent the dialogue history context in an entangled manner, which weakens the representative ability to understand the history context for expressing both the content and emotion information in the generated reply.

\section{Model}

In this section, we will firstly describe the task of empathetic response generation, and then explain the encoder and the decoder of CEDual in detail.

\subsection{Problem Statement}

Suppose in an empathetic dialogue, the dialogue history $C=\{U_1, S_1, U_2, S_2, \dots, U_t\}$ is composed of the utterances from both a speaker and a listener, where $U=\{U_1, U_2, \dots, U_t\}$ are the utterances from the speaker and $S=\{S_1, S_2, \dots, S_{t-1}\}$ are the utterances from the listener. In addition to the dialogue context, the corresponding emotion label $\bm{emo}$ is provided and represented as the one-hot vector, i.e., $\bm{emo}=[emo_1, emo_2, \dots, emo_k]$, where $k$ is the number of emotion categories, and the value corresponding to the provided emotion category is $1$. Given the dialogue history $C$ with its emotion label $emo$, the task is to understand the dialogue history and then generate the empathetic response $R$.

\subsection{Content-Emotion Duality Encoder} \label{sec:understanding}

As analyzed in Sec.~\ref{sec:intro}, the understanding of the dialogue history for empathetic reflection should be divided into the content view and the emotion view. Therefore, the Content-Emotion Duality Encoder encodes the dialogue history from two different views of content and emotion via disentanglement. 

Following \citet{lin2019moel,li2019empdg}, we encode the dialogue history to its contextual embedding $\bm{H}$ using the Transformer Encoder. To obtain the separate views of content and emotion, two different fully-connected networks are adopted to project the contextual representation $\bm{H}$ into two different spaces, i.e., the content representation $\bm{H}_c \in \mathbb{R}^{n \times d_{h}}$ and the emotion representation $\bm{H}_e \in \mathbb{R}^{n \times d_{h}}$, where $n$ is the number of the tokens in the context, and $d_h$ is the dimension of features.


While we intend to project the contextual representation $\bm{H}$ into two views using different networks, there is no guarantee 
that the content representation $\bm{H}_c$ encodes the content information only, and the emotion representation $\bm{H}_e$ encodes the emotion information only. Two disentanglement losses are designed to learn both the content-aware and emotion-aware representations based on the given emotion label $\bm{emo}$ of the dialogue history. Specifically, given the word-level content and emotion representations, we get the distributions of emotion label prediction based on their features obtained by mean-pooling. After that, we obtain the predicted results $\bm{y}_c \in \mathbb{R}^{k}$ and $\bm{y}_e \in \mathbb{R}^{k}$ based on the content and emotion features, respectively.


As mentioned in Sec.~\ref{sec:intro}, the content component of the dialogue history is the incident devoid of any feelings and may evoke different emotions. Therefore, the content feature $\bm{v}_c$ is required to be not discriminative for emotion classification. Inspired by \citet{fu2018style}, we attempt to maximize the entropy of the prediction based on the content feature $\bm{v}_c$.
\begin{equation}
    l_{dis\_c} = -\sum_{i=1}^ky_c^i{\rm log}y_c^i
\end{equation}

On the other hand, the emotion feature $\bm{v}_e$ should be discriminative enough for emotion classification based on the dialogue history. Therefore, we adopt the cross-entropy to make the emotion representation $\bm{H}_e$ encode the emotion information of the dialogue history.
\begin{equation}
    l_{dis\_e} = -\sum_{i=1}^kemo_i{\rm log}y_e^i
\end{equation}

Finally, the disentanglement loss to minimize is:
\begin{equation}
    l_{dis} = -l_{dis\_c} + l_{dis\_e}
\end{equation}


\subsection{Content-Emotion Duality Decoder} \label{sec:generation}

To exploit the content and emotion information of the dialogue history obtained by the Content-Emotion Duality Encoder, we propose two methods, namely, the first-content-then-emotion method (CEDual-FCTE) and the first-emotion-then-content method (CEDual-FETC), to generate the response step by step.

With the first-content-then-emotion method (CEDual-FCTE), the decoder first learns to generate an intermediate representation by considering the content information of the dialogue history alone and then injects the emotion information to the intermediate representation to derive an integral representation for generation by adding the emotion representation of the dialogue history. Specifically, we first obtain the output embedding $\bm{E}^R \in \mathbb{R}^{d_{emb} \times m}$ converted by the target sequence shifted by one, where $m$ is the length of the target sequence shifted by one, and $d_{emb}$ is the dimension of embeddings. Given the output embedding $\bm{E}^R$ and the content representation $\bm{H}_c$, we adopt the Transformer decoder to get the content-aware response representation, i.e.,
\begin{equation}
    \bm{V}_{fcte}^1 = TRS_{Dec}^{fcte1}(\bm{H}_c,\bm{E}^R)
\end{equation}
where $TRS_{Dec}^{fcte1}(\cdot)$ is the Transformer Decoder of the first step in the first-content-then-emotion generation process, and $\bm{V}_{fcte}^1 \in \mathbb{R}^{d_{h} \times m}$ is the temporary output of the decoder, where only the content information of the dialogue context is embedded. Then, the emotion information is introduced as follows.
\begin{equation}
    \bm{V}_{fcte}^2 = TRS_{Dec}^{fcte2}(\bm{H}_e,\bm{V}_{fcte}^1)
\end{equation}
where $\bm{V}_{fcte}^2 \in \mathbb{R}^{d_{h} \times m}$ is the emotion-enhanced response representation obtained based on the previous content-aware representation $\bm{V}_{fcte}^1$ and the emotion representation $\bm{H}_e$.

By contrast, the first-emotion-then-content method (CEDual-FETC) first obtains the emotion-aware representation and then uses the content information of the dialogue history to enhance it. Similar to CEDual-FCTE, we obtain the representation $\bm{V}_{fetc}^2$.


Using the response representation $\bm{V}_f$ (i.e. $\bm{V}_f = \bm{V}_{fcte}^2$ for first-content-then-emotion method or $\bm{V}_f = \bm{V}_{fetc}^2$ for first-emotion-then-content method), we can predict the probability distribution over the vocabulury at the current decoding step and then generate the response $R$.

To guide the training of response generation, the generation loss is designed as follows:
\begin{equation}
    l_{gen} = -{\rm log}p(R|C, \bm{emo})
\end{equation}

\subsection{Training}
As a whole, for the training purpose we minimize the sum of the disentanglement loss and the generation loss.
\begin{equation}
    l = l_{gen} + l_{dis}
\end{equation}

\section{Experiment}

\begin{table*}[t]
\centering
\begin{tabular}{l|lll|lll}
\hline
 & Acc & BLEU & Perp & Empathy & Relevance & Fluency \\ \hline
Transformer & - & 2.98 & 33.91 & 3.09 & 2.81 & 4.28 \\
EmoPrepend & 0.3328 & 3.08 & 33.35 & 3.01 & 2.66 & 4.14 \\
MoEL & 0.3200 & 2.21 & 33.58 & 3.15 & 2.87 & 4.22 \\
EmpDG & 0.3431 & 3.15 & 34.18 & 2.86 & 2.83 & 4.24 \\ \hline
CEDual-FCTE & 0.3660 & \textbf{3.50} & 32.92 & \textbf{3.31} & 2.83 & 4.37 \\
CEDual-FETC & \textbf{0.3671} & 3.32 & \textbf{32.88} & 3.26 & \textbf{2.91} & \textbf{4.39} \\ \hline
\end{tabular}
\caption{Experimental results of comparison to baselines.}
\label{tab:main results on li}
\end{table*}

\subsection{Setup}

\textbf{Datasets.}
To examine the effectiveness of our proposed model, we experiment on the dataset EMPATHETICDIALOGUES \cite{rashkin2018towards} preprocessed by \citet{li2019empdg}. The dataset consists of 25k one-to-one open-domain conversations grounded on emotional situations and provides 32 evenly distributed emotion labels.
There are 20,724 dialogues in the training set, 2,972 in the validation set, and 2,713 in the test set.

\textbf{Metrics.}
For automatic evaluation, we use BLEU \cite{papineni2002bleu}, Perplexity \cite{serban2015hierarchical} and Emotion Accuracy. For human evaluation, we follow the previous practice to qualitatively examine model performance. Specifically, we evenly sample 128 dialogues from 32 emotion catogories and then assign three human annotators to score the predicted responses generated by our proposed model as well as the compared baselines in terms of the following three metrics: Empathy, Relevance, and Fluency \cite{rashkin2018towards}. 

\textbf{Model Settings.} 
All common settings are the same as the work in \citet{lin2019moel,li2019empdg}.

\textbf{Baselines.} We compare our model with Transformer \cite{vaswani2017attention}, EmoPrepend \cite{rashkin2018towards}, MoEL \cite{lin2019moel} and EmpDG \cite{li2019empdg}.

\subsection{Comparison to Baselines}

Comparative experiment results are shown in Table.~\ref{tab:main results on li}. We observe that our proposed framework with two different decoders outperforms previous methods on both automatic and human metrics.

Specifically, CEDual-FCTE and CEDual-FETC improve the BLEU score significantly by 0.35\% and 0.17\%, and also achieve 0.43 and 0.47 decrease of Perplexity. It means that CEDual is able to generate responses with higher quality and empathy. Furthermore, the emotion accuracy is also improved by 2.29\% and 2.4\% with CEDual-FCTE and CEDual-FETC, which shows that introducing content-emotion duality is helpful for better understanding the emotion expressed by the speaker.

As for human evaluation, our model also gains promising results. Compared to MoEL, which is the best baseline, CEDual-FCTE achieves better performance on Empathy and Fluency by 0.16 and 0.15, and slightly lower performance on Relevance by 0.04. On the other hand, CEDual-FETC also improves the performance on three human metrics by 0.11, 0.04, and 0.17 compared to MoEL. The above results further verify that our model can generate better responses than previous methods from the aspect of Empathy, Relevance, and Fluency.

Besides, we also have the following findings from the experimental results. Firstly, the second step of the decoder has more influence on response generation. In fact, CEDual-FCTE achieves higher Empathy, while CEDual-FETC has better Relevance. Secondly, since the gold responses in the EMPATHETICDIALOGUES dataset often contain emotion information alone, e.g., ``I am sorry to hear that'' where the content information is missing, it makes modeling training difficult to learn more content information for generation. Consequently, CEDual-FCTE, which considers emotion information more for generation, achieves a better BLEU score and makes the responses more empathetic. In addition, the Relevance metric is more difficult to improve compared to the Empathy metric.

\subsection{Human A/B Test}

\begin{table}[t]
\centering
\resizebox {\columnwidth} {!} {
\begin{tabular}{l|l|l|l}
\hline
 & Win & Loss & Tie \\ \hline
CEDual-FCTE vs. Transformer & 0.547 & 0.398 & 0.055 \\
CEDual-FCTE vs. EmoPrepend & 0.555 & 0.351 & 0.094 \\
CEDual-FCTE vs. MoEL & 0.516 & 0.445 & 0.039 \\
CEDual-FCTE vs. EmpDG & 0.563 & 0.367 & 0.070 \\ \hline
CEDual-FETC vs. Transformer & 0.516 & 0.398 & 0.086 \\
CEDual-FETC vs. EmoPrepend & 0.555 & 0.344 & 0.101 \\
CEDual-FETC vs. MoEL & 0.523 & 0.398 & 0.078 \\
CEDual-FETC vs. EmpDG & 0.531 & 0.399 & 0.070 \\ \hline
\end{tabular}
}
\caption{Human A/B test.}
\label{tab:ab test}
\end{table}

To further illustrate whether our model outperforms the baselines, we conduct human A/B tests following \citet{lin2019moel,li2019empdg}. The results of pairwise response comparison are shown in Table.~\ref{tab:ab test}. It is observed that both CEDual-FCTE and CEDual-FETC can generate more empathetic responses than previous methods. Specifically, annotators choose more of the responses generated by CEDual-FCTE/CEDual-FETC as the more empathetic responses than Transformer, EmoPrepend, MoEL, and EmpDG by 14.8\%/11.8\%, 20.4\%/21.1\%, 7.1\%/12.5\%, and 19.6\%/13.2\%, respectively. In sum, The above results show the superiority of the CEDual.

\subsection{Ablation Study}
\label{sec:appendix}

\begin{table}[t]
\centering
\begin{tabular}{l|lll}
\hline
 & Acc & BLEU & Perp \\ \hline
CEDual-C & 0.3524 & 3.20 & 32.70 \\ \hline
CEDual-E & 0.3579 & 3.10 & 33.98 \\ \hline
CEDual-FCTE & 0.3660 & 3.50 & 32.92 \\ \hline
CEDual-FETC & 0.3671 & 3.32 & 32.88 \\ \hline
\end{tabular}
\caption{Ablation study.}
\label{tab:ablation study}
\end{table}

To further examine the effectiveness of considering the Content-Emotion Duality for generation, we conduct the following ablation tests and the experimental results are shown in Table.~\ref{tab:ablation study}.
\begin{itemize}
    \item \textbf{CEDual-C:} Only the content information is fed into the decoder.
    \item \textbf{CEDual-E:} Only the emotion information is fed into the decoder.
\end{itemize}

From the results, it is observed that if we only consider the content or emotion information for the generation, the model cannot generate more empathetic responses compared to the CEDual-FCTE and CEDual-FETC. Specifically, both the CEDual-C and CEDual-E model generate the responses with worse BLEU scores. Moreover, the two ablation models also decrease the emotion accuracy. Therefore, the ablation study shows the effectiveness of considering the Content-Emotion Duality for empathetic response generation.

\section{Conclusion}

To solve the task of empathetic response generation, in this paper we propose a Content-Emotion Duality Model, which attempts to understand the dialogue context and generate the empathetic response from both the content view and the emotion view via disentanglement.
CEDual is the first method that introduces the concept of content-emotion duality for empathetic response generation and adopts disentanglement to model content-emotion duality of a empathetic conversation. The extensive experiments verify the effectiveness of the model.

\section*{Acknowledgement}

The work described in this paper was supported by Research Grants Council of Hong Kong (PolyU 152040/18E, PolyU 15207920), National Natural Science Foundation of China (62076212) and PolyU (ZVVX).

\bibliography{anthology,custom}
\bibliographystyle{acl_natbib}

\end{document}